\documentclass[letterpaper]{article}
\usepackage{aaai}
\usepackage{times}
\usepackage{helvet}
\usepackage{courier}

\usepackage{graphicx}
\usepackage{caption}

\newtheorem{definition}{Definition}

\title{Language understanding as a step towards human level intelligence - automatizing the construction of the initial dictionary from example sentences}

\author{Chitta Baral\\
   School of Computing, Informatics and DSE\\
    Arizona State University\\ 
  \texttt{chitta@asu.edu} \\\And
 Juraj Dzifcak\\
 School of Computing, Informatics and DSE\\
Arizona State University\\ 
       \texttt{juraj.dzifcak@asu.edu} }

\date{}

\begin{document}
\maketitle

\begin{abstract}

For a system to understand natural language, it needs to be able to take natural language text and answer questions given in natural language with respect to that text; it also needs to be able to follow instructions given in natural language. To achieve this, a system must be able to process natural language and be able to capture the knowledge within that text. Thus it needs to be able to translate natural language text into a formal language. We discuss our approach to do this, where the translation is achieved by composing the meaning of words in a sentence. Our initial approach uses an inverse lambda method that we developed (and other methods) to learn meaning of words from meaning of sentences and an initial lexicon. We then present an improved method where the initial lexicon is also learned by analyzing the training sentence and meaning pairs. We evaluate our methods and compare them with other existing methods on a corpora of database querying and robot command and control. 
\end{abstract}

\section{Introduction and Motivation}

We consider natural language understanding as an important aspect of human level intelligence. But what do we mean by ``language understanding''.  In our view a system that understands language can among other attributes (i) take natural language text and then answer questions given in natural language with respect to that text and (ii) take natural language instructions and execute those instructions as a human would do.
 
A system that can do the above must have several functional capabilities, such as: (a) It must be able to process language; (b) It must be able to capture knowledge expressed in the text; (c) It must be able to reason, plan and in general do problem solving and for that it may need to do efficient searching of solutions; (d) It must be able to do high level execution and control as per given directives and (e) To scale, it must be able to learn new language aspects (for e.g., new words). These functional capabilities are often compartmentalized to different AI research topics. However, good progress in each of these areas  (over the last few decades) provides an opportunity to use results and systems from them and build up on that to develop a natural language understanding system.

Over the last two decades our group has been focusing in the research of developing suitable knowledge representation languages. The research by a broader community has led to KR languages and systems that allow us to represent various kinds of knowledge and the KR systems allow us to reason, plan and do declarative problem solving using them. Various search techniques are embedded in some of these systems and one such system from Potsdam (CLASP)\footnote{http://www.cs.uni-potsdam.de/clasp/} has been doing very well in SAT competitions\footnote{http://www.satcompetition.org/}. Similarly, various languages and systems have been developed that can take directives in a formal language and use it in high level execution and control. These cover the aspects (c) and (d) mentioned above. 

In our current research we use the existing results on (c) and (d) and develop an overall architecture that addresses the aspects (a), (b) and (e)  to lead to a natural language understanding framework. 

The first key aspect of our approach and our language understanding framework is to translate natural language to appropriate formal languages. Once that is achieved we achieve (b) and then together with  the (c) and (d) components we achieve (a). The second key aspect of our approach and our language understanding framework is that we can reason and learn about how to translate new words and phrases. This allows our overall system to scale up to larger vocabularies and thus we achieve (e).

In this paper we first give a brief presentation of our system and framework which was reported in an earlier limited audience conference/workshop. We then present some original work to enhance what was done then.

\section{Translating English to Formal languages}

\begin{figure*}[!ht]
  \begin{center}
      \includegraphics[width=12cm]{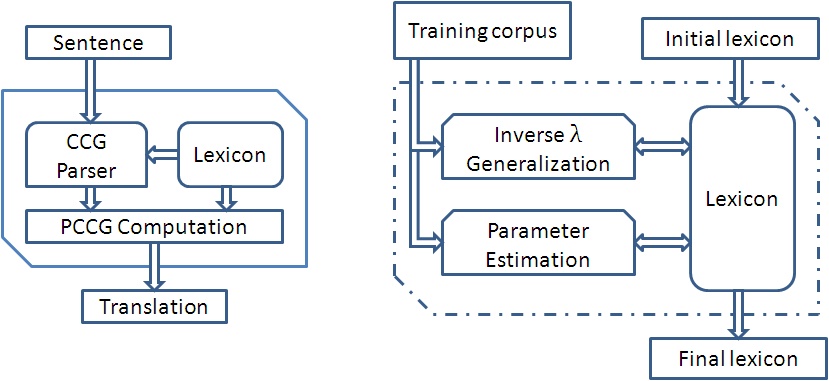}
      \caption{Overall system architecture}
      \label{figarch}
   \end{center}
\end{figure*}

Our approach to translate English to formal languages is inspired by Montague's path-breaking thesis \cite{Montague:Book} of viewing English as a formal language. We consider each word to be characterized by one or more $\lambda$-calculus formulas and the translation to be obtained by composing appropriate $\lambda$-calculus formulas of the words as dictated by a PCCG (Probabilistic Combinatorial Categorial Grammars).  The big challenge in this approach is to be able to come up with the right $\lambda$-calculus formulas for various words. Our approach, initially presented in \cite{me:iwcs}, utilizes inverse $\lambda$-calculus operators and generalization to obtain semantic representations of words and learning techniques to distinguish in between them. The system architecture of our approach is given in figure \ref{figarch}. The left block shows an overall system to translate a sentence into a target formal language using the PCCG grammar and the lexicon, while the right block shows the learning module to learn the meaning of new words (via Inverse $\lambda$ and generalization methods) and assigning weights to multiple meaning of words. We now elaborate on some important parts of the system.

\subsection{Inverse $\lambda$ computation}
The composition semantics of $\lambda$-calculus basically computes the meaning of a phrase ``a b'' by $\alpha (\beta)$ or $\beta (\alpha)$ depending on the CCG parse. Now suppose we know the meaning ``a b'' to be $\gamma$ and also know the meaning of ``a'' as $\alpha$. By inverse $\lambda$, we refer to the obtaining of $\beta$ given $\alpha$ and $\gamma$. Depending on whether $\gamma$ is $\alpha (\beta)$ or $\beta (\alpha)$ we have two inverse operators: $Inverse_R$ and $Inverse_L$. We now give a quick glimpse of $Inverse_R$ as given in \cite{me:iwcs}. Further details are given in \cite{Marcos:thesis}. 

\begin{itemize}
\item Let $G$, $H$ represent typed $\lambda$-calculus formulas, $J^1$,$J^2$,...,$J^n$ represent typed terms, $v_1$ to $v_n$, $v$ and $w$ represent variables and $\sigma_1$,...,$\sigma_n$ represent typed atomic terms.
\item Let $f()$ represent a typed atomic formula. Atomic formulas may have a different arity than the one specified and still satisfy the conditions of the algorithm if they contain the necessary typed atomic terms.
\item Typed terms that are sub terms of a typed term J are denoted as $J_i$.
\item If the formulas we are processing within the algorithm do not satisfy any of the $if$ conditions then the algorithm returns $null$.
\end{itemize}

\begin{definition}
Consider two lists of typed $\lambda$-elements A and B, $(a_i,...,a_n)$ and $(b_j,...,b_n)$ respectively and a formula $H$. The result of the operation $H(A:B)$  is obtained by replacing $a_i$ by $b_i$, for each appearance of A in H.
\end{definition}


\begin{definition}
The function $Inverse_R(H,G)$, is defined as:
\newline
\noindent Given $G$ and $H$:
\begin{it}
\begin{enumerate}
\item If $G$ is $\lambda v.v@J$, set $F = Inverse_L(H,J)$
\item If $J$ is a sub term of $H$ and G is $\lambda v.H(J:v)$
then $F$ = $J$
\item G is not $\lambda v.v@J$, $J$ is a sub term of $H$ and G is $\lambda w.H(J(J_1,...,J_m):w@J_p,...,@J_q)$ with 1 $\leq$ p,q,s $\leq$ m
then $F$ = $\lambda v_1,...,v_s.J(J_1,...,J_m:v_p,...,v_q)$.
\end{enumerate}
\end{it}
\end{definition}


To illustrate $Inverse_R$ assume that in the example given in table \ref{tab:ex2} the semantics of the word ``in'' is not known. We can use the Inverse operators to obtain it as follows. Using the semantic representation of the whole sentence, $answer(river(loc_2(stateid('arkansas'))))$, and the semantics of the word ``Name'', $\lambda x. answer(x)$, we can use the respective operators to obtain the semantics of ``the rivers in Arkansas'' as $river(loc_2(stateid('arkansas')))$. Repeating this process recursively we obtain $\lambda y. y @ loc_2(stateid('arkansas'))$ as the representation of ``in Arkansas'' and $\lambda x. \lambda y. y @ loc_2(x)$ as the desired meaning of ``in''. \footnote{A very brief review of the $\lambda$ representation is as follows. The formula $\lambda x. answer(x)$ basically means that $x$ is an input and when that input is given then it replaces $x$ in the rest of the formula. This application of a given input in expressed via the symbol $@$. Thus $\lambda x. answer(x) @ a$ reduces to $answer(a)$.}

\subsection{Generalization and trivial solution}

Using $INVERSE\_L$ and $INVERSE\_R$, we are able to obtain new semantic representations of particular words in the training sentences. To go beyond that, we use a notion of generalization that we developed. For example, consider the non-transitive verb ``fly'' who category as per a CCG\footnote{In a combinatorial categorial grammar  (CCG) words are associated with categories. The meaning of the category $S \backslash NP$ is that if a word of category $NP$ comes from the left then by combining it with a word of category $S \backslash NP$ we get a phrase of category $S$.  For example, if the word ``a'' has a category $S \backslash NP$ and the word ``b'' has category NP then the two words can be combined to the phrase ``b a'' which will have the category $S$. Similarly, the category $S / NP$ means that a word of category $NP$ has to come from the right for us to be able to combine.} \cite{Steedman:Book} is $S \backslash NP$. Lets assume we obtain a new semantic expression for ``fly'' as $\lambda x. fly(x)$ using $INVERSE\_L$ and $INVERSE\_R$. Generalization looks up all the words of the same syntactic category, $S \backslash NP$. It then identifies the part of the semantic expression in which ``fly'' is involved. In our particular case, it's the subexpression $fly$. We can then assign the expression $\lambda x. w(x)$ to the words $w$ of the same category. For example, for the verb ``swim'', we could add $\lambda x. swim(x)$ to the dictionary. This process can be performed ``en masse'', by going through the dictionary and expanding the entries of as many words as possible or ``on demand'', by looking up the words of the same categories when a semantic representation of a word in a sentence is required. Even with generalization, we might still be missing large amounts of semantics information to be able to use $INVERSE_L$ and $INVERSE_R$. To make up for this, we allow trivial solutions, where words or phrases are assigned the meaning $\lambda x. x$, $\lambda x. \lambda y. (y@x)$ or similarly simple representations, which basically mean that this word may be ignored. The trivial solutions are used as a last resort approach if neither inverse nor generalization are sufficient.

\subsection{Translation and the Overall Learning Algorithm}
Earlier we mentioned that a sentence is translated to a representation in a formal language by composing the meaning of the words in that sentences as dictated by a CCG. However, in presence of multiple meaning of words probabilistic CCG is used where the probabilities of a particular translation is computed using weights associated with each word and meaning pair. For a given sentence the translation that has the higher probability is picked. This raises the question of how does one obtain the weights. The weights are obtained using standard parameter estimation approaches with the goal that the weights should be such that they maximize the overall probability of translating each of the sentences in the training set (of sentences and their desired meaning) to their desired meaning. We now present our overall learning algorithm that combines inverse $\lambda$, generalization and parameter estimation.

\begin{tiny}
\begin{itemize}
\item {\bf Input:}  A set of training sentences with their corresponding desired representations $S = \{(S_i,L_i) : i = 1 . . . n\}$ where
$S_i$ are sentences and $L_i$ are desired expressions. Weights are given an initial value of $0.1$. An initial feature vector $\Theta_0$.

\item {\bf Output:} An updated lexicon $L_{T+1}$.  An updated feature vector $\Theta_{T+1}$.

\item {\bf Algorithm:}

\begin{itemize}

\item Set $L_0 = INITIAL\_DICTIONARY(S)$

\item For t = 1 . . . T

\item Step 1: (Lexical generation)

\item For i = 1...n.
\begin{itemize}
\item For j = 1...n.

\item Parse sentence $S_j$ to obtain $T_j$
\item Traverse $T_j$
\begin{itemize}
\item  apply $INVERSE\_L$, $INVERSE\_R$ and $GENERALIZE_D$ to find new $\lambda$-calculus expressions of words and phrases $\alpha$.
\end{itemize}

\item Set $L_{t+1} = L_t \cup \alpha$

\end{itemize}
\item Step 2: (Parameter Estimation)

\item Set $\Theta_{t+1} = UPDATE(\Theta_t, L_{t+1})$\footnote{For details on $\Theta$ computation, please see \cite{Collins:2005}}
\end{itemize}
\item return $GENERALIZE(L_T, L_T), \Theta(T)$
\end{itemize}
\end{tiny}

\section{Automatic generation of initial dictionary}

In tables 6 and 7 we compare the performance of our systems INVERSE, INVERSE+, and INVERSE+(i) with other systems that have similar goals. However, although other systems had other issues, we were not happy that our systems required a manually created initial dictionary consisting of $\lambda$-calculus representations of a set of words.  In the rest of the paper we present an approach to overcome that by automatically coming up with candidates for the initial dictionary and letting the parameter estimation module figure out the correct meaning. In particular we present methods to automatically come up with possible $\lambda$-calculus representation of nouns and various other words that are part of the initial vocabulary in \cite{me:iwcs}. Unlike \cite{me:iwcs}, where each of the word in the initial vocabulary is given a unique $\lambda$-calculus representation, our approach does not necessarily come up with a single $\lambda$-calculus representation of the words that are in the initial vocabulary in \cite{me:iwcs} but sometimes may come up with multiple possibilities.

We will now illustrate our approach in obtaining the initial dictionary and the use of CCG and $\lambda$-calculus in obtaining semantic representations of sentences on the Geoquery corpus at http://www.cs.utexas.edu/users/ml/geo.html. Table \ref{tab:ex1} shows several examples of sentences with their desired representations while table \ref{tab:ex2} shows a sample CCG parse with it's corresponding semantic derivation. 

\begin{table*}[htb]
\tiny{
\begin{center}
\begin{tabular}{|c|c|}
\hline
Sentence & Representation \\
\hline
\hline
Name the rivers in Arkansas. & $answer(river(loc_2(stateid('arkansas'))))$ \\
\hline
How many people are there in New York? & $answer(population_1(stateid('new york')))$\\
\hline
How high is Mount McKinley? & $answer(elevation_1(placeid('mount mckinley')))$\\
\hline
Name all the lakes of US. & $answer(lake(loc_2(countryid('usa'))))$\\
\hline
Name the states which have no surrounding states. & $answer(exclude(state(all), next_to_2(state(all))))$\\
\hline
\end{tabular}

\end{center}
}
\caption{Example translations.}
\label{tab:ex1}
\end{table*}


\begin{table*}[htb]
\tiny{
\begin{center}
\begin{tabular}{c c c c c}
Name & the & rivers & in & Arkansas. \\
$S/NP$ & $NP/NP$ & $N$ & $(NP\backslash N) /N$ & $N$ \\
\cline{4-5}
$S/NP$ & $NP/NP$ & $N$ & $NP\backslash N$ & \\
\cline{3-5}
$S/NP$ & $NP/NP$ & & $NP$ & \\
\cline{2-5}
$S/NP$ & $NP$ & & & \\
\cline{1-5}
$S$ & & & & \\
\end{tabular}

\begin{tabular}{c c c c c}
Name & the & rivers & in & Arkansas. \\
$\lambda x. answer(x)$ & $\lambda x. x$ & $\lambda x. river(x)$ & $\lambda x. \lambda y. y @ loc_2(x)$ & $stateid('arkansas')$ \\
\cline{4-5}
$\lambda x. answer(x)$ & $\lambda x. x$ & $\lambda x. river(x)$ & $\lambda y. y @ loc_2(stateid('arkansas'))$ & \\
\cline{3-5}
$\lambda x. answer(x)$ & $\lambda x. x$ &  & $river(loc_2(stateid('arkansas')))$ & \\
\cline{2-5}
$\lambda x. answer(x)$ & $river(loc_2(stateid('arkansas')))$ & & &\\
\cline{1-5}
$answer(river(loc_2(stateid('arkansas'))))$ & & & &\\
\end{tabular}
\end{center}
}
\caption{CCG and $\lambda$-calculus derivation for ``Name the rivers in Arkansas.''}
\label{tab:ex2}
\end{table*}

To be able to automatically create the entries in the initial dictionary as given by \cite{me:iwcs}, we need to answer the following two questions. {\it How do we find the expression $\lambda x. answer(x)$ and how do we assign it to the word ``Name''?}. The word ``answer'' isn't given anywhere by the sentence. Similarly, {\it How do we know that the semantic expression for ``Arkansas'' should be $stateid('arkansas')$?}. The first question can be answered by looking at several possible semantic representations as given in table \ref{tab:ex1}. They share one common aspect, which is that they all contain the predicate $answer$ as the outermost expression. Thus, we can assume that $\lambda x. answer(x)$ should be part of any derivation as given by table \ref{tab:ex2}. In general, using the grammar derivations for the meaning representations, we can compare various representations and look for common parts, which we will refer to as {\it common structures}. We identify these common parts and assign them to certain {\it relevant} words in the sentence, such as assigning the common expression $\lambda x. answer(x)$ to the word ``Name''. To answer the second question, we again look at the grammar derivations for nouns, and analyze them to be able to obtain the semantic expression for ``Arkansas'' as $stateid('arkansas')$.  

Table \ref{tab:ex2} shows an example syntactic and semantic derivation for the sentence ``Name the rivers in Arkansas.''. The syntactic categories for each are given by the upper part of the table. These are then combined using combinatorial rules \cite{Steedman:Book} to obtain the rest of the syntactic categories. For example, the word ``Arkansas'' of category $N$ is combined with the word ``in'' of category $(NP\backslash N) /N$, to obtain the syntactic category of ``in Arkansas'', $NP\backslash N$. The lower portion of the table lists the semantic representations of each words using $\lambda$-calculus. These are combined by applying the formulas one to another, following the syntactic parse tree. For example, the semantics of ``Arkansas'', $stateid('arkansas')$, is applied onto the semantics of ``in'',  $\lambda x. \lambda y. y @ loc_2(x)$, yielding $\lambda y. y @ loc_2(stateid('arkansas'))$. 

Let us first discuss the common structures of a logical form. For example, for the Geoquery corpus, as shown in table \ref{tab:ex1}, many queries are of the form $answer(X)$ where $X$ is a structure corresponding to the actual query. Similarly, by analyzing the Robocup corpus, we realize that all the queries are of the form $((A)$ $(do B))$ , $(definer$ $C$ $(B))$ or $(definec$ $C$ $(B))$, where $C$ is an identifier and $A$ and $B$ are some other constructs in the given language. The main attribute of these expressions is that they define the structure(s) of the desired meaning representation. 

The second component of the dictionaries were the semantic representations of nouns. Unlike the common structures, these need to be generated for as many nouns as possible to ensure that the system is capable to learn the missing semantic representations. For example, in GeoQuery, a noun ``Arkansas'' is represented as $stateid('arkansas')$. \footnote{We are using the funql representation, although the same approach is applicable for the prolog one.} For Robocup, a compound noun ``player 5'' can be represented as $(player$ $our$ $\{5\})$. 

Thus our task in being able to automatically obtain these is two fold. We first need to identify the common structures and find the appropriate $\lambda$-calculus formulas and, pick the words to which we will assign them. The second part of our goal is to find the corresponding $\lambda$-calculus expressions for nouns and compound nouns. 

We will assume this process is done on the training data and full syntactic parse of the sentences, as well as the parse of the desired formal representation are given. 

\subsubsection{Common structures}

In order to look for the common structures, we will compare the derivation structures of various formulas and look for common structures in them. To limit the potential search, and with respect to our previous experience, we will only look for the common parts at top parts of the derivation. Also, in order to be more precise and keep the computation within reasonable bounds, instead of looking at the whole grammar for meaning representations, we will look at the derivations of the meaning representations of the training data. This is a reasonable assumption, as in general the amount of structures in the target language can be assumed to be less than the amount of training data as in the case of Geoquery and CLANG.

\begin{definition}
Given a context free grammar $G$ with an initial symbol $S$, a set of non-terminals $N$, a set of terminals $T$, a set of production rules $P$ and a string $w = x_1,...,x_n$, where $x_i$s are terminal or non-terminal symbols, a production $d$ is a transformation $x_1,...,x_n \Rightarrow x_1,...,x_{i-1},A, x_{i+1},...,x_n$ such that $x_i \rightarrow A$ is in $P$. We will say that $x_i \rightarrow A$ corresponds to $d$.

Given a sequence of productions $d* = d_1,...,d_n$ a derivation tree $t$ corresponding to $d*$ is given as:

- If $n=1$, let $X \rightarrow X_1,X_2,...,X_n$ be the rule corresponding to $d_1$. Then $t$ is a tree with $X$ as the root node, which has $n$ children, in order, left to right, $X_1,X_2,...,X_n$.

- If $t'$ is a derivation tree corresponding to $d_1,...,d_{n-1}$ and $X \rightarrow X_1,X_2,...,X_n$ is the rule corresponding to $d_n$, then $t$ is given as $t'$ with $n$ children added, in order, left to right, $X_1,X_2,...,X_n$, to the left most leaf $X$ of $t'$.


A $\lambda$ tree is a pair $(V,t)$, where $V$ is a list of $\lambda$ bound variables and $t$ is a tree, where each interior node of $t$ is a non terminal symbol from $N$ and each leaf node of $t$ is a terminal symbol from $T$ or a variable from $V$.

Given two sequences of productions $d_1$ and $d_2$ with their corresponding derivation trees $t_1$ and $t_2$, a $\lambda$ tree $(V, t_c)$ is a common template of $t_1$ and $t_2$ iff there exists two sequences of applications $s_1 = X_1,...,X_n$ and $s_2 = Y_1,...,Y_n$ such that when we apply each $X_i$ to each $v_i$, $i = 1,...,n$, in $t_c$ we obtain a subtree of $t_1$ and when we apply each $Y_i$ to each $v_i$, $i = 1,...,n$, in $t_c$ we obtain a subtree of $t_2$.

\end{definition}

Example derivation trees and a common template are given in tables \ref{tab:ext} and \ref{tab:exc}.

\begin{table*}[htb]
\tiny{
\begin{center}
\begin{tabular}{c c c c c c c c c c c c c c c c c c c c}
 & & & $S$ & & & & & & $S$ & & & & & & $CITY$ & & & & \\
 & & & & & & & & & & & & & & & & & & & \\
 & & $\swarrow$ & $\downarrow$ & $\searrow$ & & & & $\swarrow$ & $\downarrow$ & $\searrow$ & & & & $\swarrow$ & $\downarrow$ & $\searrow$ & &\\
 & & & & & & & & & & & & & & & & & & & \\
 & $answer($ & & $RIVER$ & & $)$ & & $answer($ & & $PLACE$ & & $)$ & & $city($ & & $CITY$ & & $)$ & \\
 & & & & & & & & & & & & & & & & & & & \\
 & & $\swarrow$ & $\downarrow$ & $\searrow$ & & & & $\swarrow$ & $\downarrow$ & $\searrow$ & & & & $\swarrow$ & $\downarrow$ & $\searrow$ & &\\
 & & & & & & & & & & & & & & & & & & & \\
 & $river($ & & $RIVER$ & & $)$ & & $lake($ & & $PLACE$ & & $)$ & & $loc_2($ & & $STATE$ & & $)$ & \\
 & & & & & & & & & & & & & & & & & & & \\
 & & $\swarrow$ & $\downarrow$ & $\searrow$ & & & & $\swarrow$ & $\downarrow$ & $\searrow$ & & & & $\swarrow$ & $\downarrow$ & $\searrow$ & &\\
 & & & & & & & & & & & & & & & & & & & \\
 & $loc_2($ & & $STATE$ & & $)$ & & $loc_2($ & & $COUNTRY$ & & $)$ & & $stateid($ & & $STATENAME$ & & $)$ & \\
 & & & & & & & & & & & & & & & & & & & \\
 & & & & & & & & & & & & & & & $\downarrow$ & & &\\
 & & & & & & & & & & & & & & & & & & & \\
 & & & & & & & & & & & & & & & $'virginia'$ & & &\\
\end{tabular}
\end{center}
}
\caption{Sample derivation trees}
\label{tab:ext}
\end{table*}

\begin{minipage}[b]{.40\textwidth}
\centering
\small{
\begin{tabular}{c c c c c c c c}
 & & & & $S$ & & & \\
 & & & & & & & \\
 $\lambda v.$& & & $\swarrow$ & $\downarrow$ & $\searrow$ & &\\
 & & & & & & & \\
 & & $answer($ & & $v$ & & $)$ & \\
\end{tabular}
}
\captionof{table}{Sample common template.}
\label{tab:exc}
\end{minipage}\qquad

Thus, based on the above definitions, to look for the common structures in the desired meaning representations, we will look for common trees between derivations which are rooted at the initial symbol. As an example, consider the following parts of the derivation, obtained directly from the Geoquery corpus, for $answer(river(loc_2(stateid('arkansas'))))$ and $answer(lake(loc_2(countryid('usa'))))$.

%
\begin{center}
\begin{tiny}
\begin{itemize}
\item $S \rightarrow_{(1a)} answer( RIVER )$ 
\item $\rightarrow_{(2a)} answer(river ( RIVER ))$ 
\item $\rightarrow_{(3a)} answer(river (loc_2 ( STATE )))$ 
\end{itemize}
\end{tiny}
\end{center}

\begin{center}
\begin{tiny}
\begin{itemize}
\item $S \rightarrow_{(1b)} answer( PLACE )$
\item $\rightarrow_{(2b)} answer( lake ( PLACE ))$
\item $\rightarrow_{(3b)} answer( lake (loc_2 ( COUNTRY )))$
\end{itemize}
\end{tiny}
\end{center}

Starting from the initial non-terminal $S$, we can see that the rules (1a) and (1b) are already different. They share a common part in having the terminal symbols $answer($ and $)$. Thus, if we replace all the non-terminals in the common parts of the derivation with $\lambda$ bound variables, we obtain the common part of the derivations as $\lambda v. answer(v)$, where $v$ is the new $\lambda$ bound variable.

In general, having a derivation, we start at the initial symbol and follow the derivation tree level by level while comparing the nodes in the derivation tree. We then collect all the common terminals from this subtree, and replace all the different non-terminals with $\lambda$ bound variables. Note that there might be multiple such structures, as in the case of Robocup corpus. In that case we would store and use all of them and the learning part of the system would take care of picking the proper ones. 

After finding the common structures between the derivations, we need to find the words to which we assign them to. Since the structures are supposed to define the common structures of the desired representations, it is reasonable to try to assign them to words which, in a sense, ``define'' the sentences. In our case, we look for words that are usually last to combine in the CCG derivation. The reasoning is that when looking for the common structures, we looked at the top parts of the derivation of meaning representations. Thus it is reasonable to try to assign them to words which are in the top parts of the derivation in the syntactic parse of the sentence. Note that these words might not be the ones with most complex categories. In practice, such words are usually verbs, wh-words or some adverbs.

\begin{definition}
Given a CCG parse tree $T$ of a sentence $s$ and a word $w$ from $s$, a word $w$ is a {\it top} word if there is no other word $w'$ from $s$, such that  $level(w') < level(w)$.
\end{definition}

Given a set of training pairs $(S_i, L_i)$, $i = 1,...,k $, where $S_i$ is a sentence and $L_i$ is the corresponding desired logical form, together with a syntactic parse of $S_i$ and the derivation of $L_i$, we can obtain the candidate common structures using the following algorithm, denoted as $INITIAL_C$.

\begin{tiny}
\begin{itemize}
\item {\bf Input:}

A set of training sentences with their corresponding desired representations $S = \{(S_i,L_i) : i = 1 . . . n\}$ where
$S_i$ are sentences and $L_i$ are desired expressions. 
A CCG grammar G for sentences $S_i$. A CFG grammar G' for representations $L_i$.

\item {\bf Output:}

An initial lexicon $L_0$.

\item {\bf Algorithm:}

\begin{itemize}

\item Step 1: (Word selection)

\item For i = 1...n.
\begin{itemize}
\item	Parse $S_i$ using the CCG grammar G to obtain parse tree $t_i$. Find all the top words of $t_i$ and store them in $W_i$.
\end{itemize}

\item Step 2: ($\lambda$-expression generation)

\item For i = 1...n.
\begin{itemize}
\item For j = 1...n.

\item Parse derivations $L_i$ and $L_j$ using the CFG grammar G' to obtain the derivation trees $T_i$ and $T_j$.
\item Starting from roots, compare $T_i$ and $T_j$ and find the largest common template $(V,T)$, such that $T$ that is rooted at the initial symbol of the grammar, $S$.
\item Concatenate all the leafs of $T$ together to form a $\lambda$-expression $\gamma$. For each $v \in V$, add $\lambda v.$ in front of $\gamma$.
\item Add $\gamma$ as semantic expression to each of the words in $W_i$ and $W_j$.\footnote{This step exhaustively assigns the new semantics to all the top words. While not optimal, the learning part of the overall algorithm takes care of figuring out the proper assignment.}
\end{itemize}

\item Set $L_0 = \cup_{i} W_i$ 
\item return $L_0$
\end{itemize}
\end{itemize}
\end{tiny}

\subsubsection{Nouns}

In order to derive potential $\lambda$-expression candidates for nouns, instead of looking at the top of the derivation trees and finding words, we match the nouns with the terminals in the leafs of the derivation tree and then look for non-terminals which can produce it. As we  traverse upwards towards the root, we look for other terminals which are produced by the non-terminals we encounter. At each encountered non-terminal, we generate potential candidate $\lambda$-expressions by analyzing the current subtree and store them. As in the previous case, we leave it to the parameter learning part of the overall algorithm to figure out the proper ones. Our approach can be illustrated as follows.

Let us look at an example of rules deriving $(city(loc_2(stateid('virginia'))))$ from the sentence ``Give me the cities in Virginia.'', also given by table \ref{tab:ext}.

\begin{center}
\begin{tiny}
\begin{itemize}
\item $CITY \rightarrow_{1f} city ( CITY )$ 
\item $CITY \rightarrow_{2f} loc_2 ( STATE )$ 
\item $STATE \rightarrow_{3f}  stateid ( STATENAME )$ 
\item $STATENAME \rightarrow_{4f} \hspace{0 pc} 'virginia'$ 
\end{itemize}
\end{tiny}
\end{center} 

Let us assume that the noun we are interested in is ``Virginia''. First, we will attempt to match it to a terminal in the derivation, which in this case is $' virginia '$. We will then traverse the tree upwards. In this case, we first reach the non-terminal $STATENAME$. Since $' virginia '$ is the only child, we add $' virginia '$ as the potential candidate representation of ``Virginia''. Continuing recursively, we arrive at the non-terminal $STATE$. It has additional terminal symbols as children, $stateid($ and $)$. We try to match these with the sentence and after being unsuccessful, we concatenate on the leafs of the current subtree to generate another potential candidate, which yields $stateid (' virginia ')$. Continuing to traverse we arrive at the non-terminal $CITY$ in the rule $(2f)$. As in the previous case, it has terminal symbols $loc_2($ and $)$ as children, and we are unable to match them onto the sentence. Thus we again concatenate at the leaves, leading to $loc_2(stateid (' virginia '))$ as a potential representation candidate for the word ``Virginia''. Continuing upwards in the tree, we reach the non-terminal symbol $CITY$ given by the rule $(1f)$. In this case, we can match one of it's children, the terminal $city ($, with some words in the sentence and we stop. This approach produces three possible representations for ``Virginia'', $' virginia '$, $stateid (' virginia ')$, $loc_2(stateid (' virginia '))$. However, during the training process the first one does not yield any new semantic data using the inverse lambda operators, while the third one is too specific and can only be used in very few sentences. Consequently, their weights are very low and they are not used, leaving $stateid (' virginia ')$ as the relevant representation.

We will now define an algorithm to obtain the candidate noun expressions from the training set, denoted by $INITIAL_N$. For our experiments, $maxlevel$ was set to $2$ and $accuracy$ was set to $0.7$. 


\begin{tiny}
\begin{itemize}
\item {\bf Input:} A set of training sentences with their corresponding desired representations $S = \{(S_i,L_i) : i = 1 . . . n\}$ where
$S_i$ are sentences and $L_i$ are desired expressions. 
A CCG grammar G for sentences $S_i$. A CFG grammar G' for representations $L_i$.

$FN(t)$ - given a CCG parse tree $t$, returns all the nouns in $t$
$nMATCH(w)$ - returns a set of terminal symbols partially matching the string $w$ with accuracy $a$. Returns a single non terminal if $w$ is a single word. The accuracy for $nMATCH(w)$ is given by the partial string matching, given as the percentage of similar parts in between the strings
$MCYK(X)$ - given a set of terminal and non terminal symbols, finds the non-terminal symbol which can yield all of them using a modified CYK algorithm.

$maxlevel$ $M$ - maximum number of levels allowed to traverse in the derivation trees

\item {\bf Output:} An initial lexicon $L'_0$.

\item {\bf Algorithm:}

\item Step 1: ($\lambda$-expression generation)

\item For i = 1...n.
\begin{itemize}
\item	Parse $S_i$ using the CCG grammar to obtain $t_i$.
\item	Parse $L_i$ using the CFG grammar to obtain $T_i$.
\item Set $W = FN(t_i)$.
\item For each $w_j \in W$:
\begin{itemize}
\item Set $X = nMATCH(w)$
\item Repeat a maximum of $M$ times
\begin{itemize}
\item Set $N$ = $MCYK(X)$.
\item Set $T$ to be a subtree of $T_i$ rooted at the $N$. 
\item For each leaf node $n$ of $T$ which is a match of some word $w'$ of the sentence $S_i$, if the path from $n$ to $N$ contains a non-terminal symbol, replace $n$ with a new $\lambda$ bound variable $v$ and add $\lambda v.$ to $\Gamma$
\item Concatenate all the leaf nodes of $T$ to form $\Gamma'$.
\item Set $\Gamma$ = $\Gamma$. $\Gamma'$, where '.' represents string concatenation
\item Add $(w_j, \Gamma)$ to $L'_0$.
\item If $N$ has two or more non-terminal children, break.
\item If $N$ has a child which terminal symbol can be matched to any word of $S_i$ but $w_j$, break.
\item Set $N$ = $MCYK(N)$.
\end{itemize}
\end{itemize}

\end{itemize}
\item return $L'_0$.
\end{itemize}
\end{tiny}

The algorithm stops when it encounters other terminals because we are looking for the representations of specific words. We assume each word is represented as a lambda calculus formula. Once we encounter a terminal corresponding to some other word of the sentence, we assume that word has it's own representation which we do not want to add to the representation of the current noun we are investigating. The algorithm produces results such as $\lambda x, answer(x)$ for the words $list$, $name$, $what$ and $stateid('virginia')$ for the word $Virginia$. In case of CLANG corpus, some of the results are $\lambda x. \lambda y. (x) (do$ $y)$, $\lambda x. \lambda y. definer$ $'x'$ $y$ for each of the words $call$, $let$, $if$.  

Combining the output of both algorithms yields an initial lexicon which can be used by the system. Some of the results obtained by the algorithms are given in table \ref{ex-res}.

\begin{minipage}[b]{.40\textwidth}
\centering
\begin{tiny}
\begin{tabular}{|c| c|}
\hline
 Word & Obtained representations \\
\hline
\hline
list & $\lambda x, answer(x)$ \\
\hline
Virgina & $stateid('virginia')$ \\
\hline
what & $\lambda x, answer(x)$ \\
\hline
Mississippi & $stateid ('mississippi')$, $riverid ('mississippi')$ \\
\hline
if & $\lambda x. \lambda y. (x) (do$ $y)$ \\
  & $\lambda x. \lambda y. definer$ $'x'$ $y$ \\
\hline
let & $\lambda x. \lambda y. (x) (do$ $y)$ \\
  & $\lambda x. \lambda y. definer$ $'x'$ $y$ \\
\hline
player 5 & $(player$ $our$ $\{5\})$ \\
\hline
midfield & $\lambda x. (x$ $midfield)$ \\
\hline
\end{tabular}
\end{tiny}
\captionof{table}{Examples of learned initial representations.}
\label{ex-res}
\end{minipage}\qquad

\section{Evaluation}


Similarly to \cite{Collins:2009}, we used the standard GEOQUERY and CLANG corpora for evaluation.
The GEOQUERY corpus contained 880 English sentences with their respective database queries in $funql$ language. The CLANG corpus contained 300 entries specifying rules, conditions and definitions in CLANG. 

In all the experiments, we used the $C\&C$ parser of \cite{CCG} to obtain syntactic parses for sentences. In case of CLANG, most compound nouns including numbers were pre-processed. We used the standard 10 fold cross validation and proceeded as follows. A set of training and testing examples was generated from the respective corpus. These were parsed using the $C\&C$ parser to obtain the syntactic tree structure. Next, the syntactic parses plus the grammar derivations of the desired representations for the training data were used to create a corresponding initial dictionary. These together with the training sets containing the training sentences with their corresponding semantic representations (SRs) were used to train a new dictionary with corresponding parameters. Note that it is possible that many of the words were still missing their SRs, however note that our generalization approach was also applied when computing the meanings of the test data. This dictionary was then used to parse the test sentences and the highest scoring parse was used to determine precision and recall. Since many words might have been missing their SRs, the system might not have returned a proper complete semantic parse. To measure precision and recall, we adopted the measures given by \cite{Mooney:2007} and \cite{Mooney:2009}. {\it Precision} denotes the percentage of of returned SRs that were correct, while {\it Recall} denotes the percentage of test examples with pre-specified SRs returned. {\it F-measure} is the standard harmonic mean of precision and recall. For database querying, a SR was correct if it retrieved the same answer as the standard query. For CLANG, an SR was correct if it was an exact match of the desired SR, except for argument ordering of conjunctions and other commutative predicates. 


To evaluate our system, a comparison with the performance results of several alternative systems with available data is given. In many cases, the performance data given by \cite{Mooney:2009} are used. We compared our system with the following ones: The SYN0, SYN20 and GOLDSYN systems by \cite{Mooney:2009}, the system SCISSOR by \cite{Mooney:2005}, an SVM based system KRIPS by \cite{Mooney:2006}, a synchronous grammar based system WASP by \cite{Mooney:2007}, the CCG based system by \cite{Collins:2007}, the work by \cite{Lu:2008} and the INVERSE and INVERSE+ systems given by \cite{me:iwcs}. The results for different copora, if available, are given by the tables \ref{table-resGeo} and \ref{table-resClang}\footnote{ The $INVERSE+(i)$ and $A-INVERSE+(i)$ denotes evaluation where ``(definec'' and ``(definer'' at the start of SRs were treated as being equal.}. The work by \cite{liang:2011} reports a $91.1\%$ recall on geoquery corpus but uses a 600 to 280 split.


\begin{minipage}[b]{.40\textwidth}
\centering
\begin{tiny}
\begin{tabular}{|c| c| c| c|}
\hline
 & Precision & Recall & F-measure \\
\hline
A-INVERSE+  &  94.58 & 90.22 & 92.35\\
\hline
INVERSE+  &  93.41 & 89.04 & 91.17\\
INVERSE   &  91.12 & 85.78 & 88.37\\
\hline
GOLDSYN   &  91.94 & 88.18 & 90.02\\
\hline
WASP      &  91.95 & 86.59 & 89.19\\
\hline
Z\&C      &  91.63 & 86.07 & 88.76\\
\hline
SCISSOR   &  95.50 & 77.20 & 85.38\\
KRISP     &  93.34 & 71.70 & 81.10\\
Lu at al. &  89.30 & 81.50 & 85.20\\
\hline
\end{tabular}
\end{tiny}
\captionof{table}{Performance on GEOQUERY.}
\label{table-resGeo}
\end{minipage}\qquad
\begin{minipage}[b]{.40\textwidth}
\centering
\begin{tiny}
\begin{tabular}{|c| c| c| c|}
\hline
 & Precision & Recall & F-measure \\
\hline
A-INVERSE+  &  87.05 & 79.28 & 82.98\\
\hline
INVERSE+(i)&  87.67 & 79.08 & 83.15\\
INVERSE+   &  85.74 & 76.63 & 80.92\\
\hline
GOLDSYN   &  84.73 & 74.00 & 79.00\\
SYN20     &  85.37 & 70.00 & 76.92\\
SYN0      &  87.01 & 67.00 & 75.71\\
\hline
WASP      &  88.85 & 61.93 & 72.99\\
KRISP     &  85.20 & 61.85 & 71.67\\
\hline
SCISSOR   &  89.50 & 73.70 & 80.80\\
Lu at al. &  82.50 & 67.70 & 74.40\\
\hline
\end{tabular}
\end{tiny}
\captionof{table}{Performance on CLANG.}
\label{table-resClang}
\end{minipage}\qquad




The results of our experiments indicate that our approach outperforms the existing parsers in F-measure and illustrate that our approach scales well and is applicable for sentences with various lengths. In particular, it is even capable of outperforming the manually created initial dictionaries given by \cite{me:iwcs}. The main reason seems to be that unlike in \cite{Mooney:2007}, our approach actually benefits from a more simplified nature of funql compared to PROLOG. The resulting $\lambda$-calculus expressions are often simpler, as they do not have to account for variables and multiple predicates. The increase in accuracy mainly resulted from the decrease of number of possible semantic expressions of words. As we understand the work by \cite{me:iwcs} would sometimes include many meanings of words. Our approach reduces this number. A decrease was caused by not being able to automatically generate some expressions that were manually added in Baral et al 2011. The automatically obtained dictionary contained around 32\% of the semantic data of the manually created one.

Most of the failures of our system can be attributed to the lack of data in the training set. In particular, new syntactic categories, or semantic constructs rarely seen in the training set usually result in complete inability to parse those sentences. In addition, given the syntactic parses, a complex semantic representations in lambda calculus are produced, which are then often propagated via generalization and can produce bad translation and interfere with learning. Additionally, many of the words will have several possible representations and the training set distribution might not properly represent the desired one. The $C\&C$ parser that we used was primarily trained on news paper text, \cite{CCG}, and thus did have some problems with these different domains and in some cases resulted in complex semantic representations of words. This could be improved by using a different parser, or by simply adjusting some of the parse trees. 

In the previous paragraphs we compared our system with similar systems in terms of performance. We now give a qualitative comparison of our approach with other learning based approaches that can potentially translate natural language text to formal representation languages \cite{Collins:2005}, \cite{Mooney:2006}, \cite{Mooney:2006a}, \cite{Mooney:2007}, \cite{Lu:2008}, \cite{Collins:2007}, \cite{Mooney:2009}, \cite{Zettlemoyer:2010}, \cite{Zettlemoyer:2011}, \cite{liang:2011}. \cite{Collins:2005} uses a set of hand crafted rules to learn syntactic categories and semantic representations of words using combinatorial categorial grammar (CCG), \cite{Steedman:Book}, and $\lambda$-calculus formulas, \cite{ltfgamut91}. The same approach is adopted in \cite{Collins:2007}. \cite{Kanazawa:2001}, \cite{Kanazawa:2003} and \cite{Kanazawa:2006} focuses on computing the missing $\lambda$-expressions, but do not provide a complete system. In \cite{Mooney:2009}, a word alignment approach is adopted to obtain the semantic lexicon and rules, which allow semantic composition, are learned. Compared to \cite{Mooney:2009}, we do not generate word alignments for the sentences and their semantic representations. We only use a limited form of pattern matching to initialize our approach with several basic semantic representations. We focus on the simplest cases, the top and bottom of the trees, rather than performing a complete analysis of the trees. We assign each word a $\lambda$-calculus formula as it's semantics and use the native $\lambda$-calculus application, $@$, to combine them rather than computed composition rules. The learning process then figures out which of the candidate semantics to use. We use a different syntactic parser which dictates the direction of the semantic composition. Both approaches use a similar learning model based on \cite{Collins:2005}. The work by \cite{Zettlemoyer:2010} uses higher-order unification. Instead of using inverse, they perform a split operation which can break a $\lambda$ expression into two. However, this approach is not capable of learning more complex $\lambda$ calculus formulas and lacks generalization. \cite{liang:2011} uses dependency-based compositional semantics(DCS) with lexical triggers which loosely correspond to our initial dictionaries.  

\section{Conclusion and Discussion}

In this work we presented an approach to translate natural language sentences into semantic representations. Using a training set of sentences with their desired semantic representations our system is capable of learning the meaning representations of words. It uses the parse of desired semantic representations under an unambiguous grammar to obtain an initial dictionary, inverse $\lambda$ operators and generalization techniques to automatically compute the semantic representations based on the syntactic structure of the syntactic parse tree and known semantic representations without any human supervision. Statistical learning approaches are used to distinguish the various potential semantic representations of words and prefer the most promising one. In this work, we are able to overcome some of the deficiencies of our initial work in 
 \cite{me:iwcs}.  Our approach here is fully automatic and it generates a set of potential candidate words for each noun based solely on the context free grammar of the target language and the training data. The resulting method is capable of outperforming many of the existing systems on the standard copora of Geoquery and CLANG. There are many possible extensions to our work. One of the possible direction is to experiment with additional corpora which uses temporal logic as a target language. Other directions include the improvements in inverse lambda computation and application of other learning methods such as sparse learning.

\bibliographystyle{aaai}
\bibliography{bib-all}

\end{document}